# Comparison Analysis of Facebook's Prophet, Amazon's DeepAR+ and CNN-QR Algorithms for Successful Real-World Sales Forecasting

Emir Žunić[1,2], Kemal Korjenić[1], Sead Delalić[3,1] and Zlatko Šubara[1]

[1]Info Studio d.o.o. Sarajevo, Bosnia and Herzegovina
[2]Faculty of Electrical Engineering, University of Sarajevo, Bosnia and Herzegovina
[3]Faculty of Science, University of Sarajevo, Bosnia and Herzegovina

*Abstract*

*By successfully solving the problem of forecasting, the processes in the work of various companies are optimized and savings are achieved. In this process, the analysis of time series data is of particular importance. Since the creation of Facebook's Prophet, and Amazon's DeepAR+ and CNN-QR forecasting models, algorithms have attracted a great deal of attention. The paper presents the application and comparison of the above algorithms for sales forecasting in distribution companies. A detailed comparison of the performance of algorithms over real data with different lengths of sales history was made. The results show that Prophet gives better results for items with a longer history and frequent sales, while Amazon's algorithms show superiority for items without a long history and items that are rarely sold.*

*Keywords*

*Sales forecasting, Real-world dataset, Prophet, DeepAR+, CNN-QR, Backtesting, Classification*

## 1. Introduction

Successful sales forecasting mechanisms can have positive effects in many areas of business, and one of the basic aspects is stock optimization. In retail, wholesale and distribution companies, inventory optimization is one of the key aspects of business. Companies that maintain their stocks at an adequate and satisfactory level can save significant amounts of money, and at the same time their other processes, such as warehousing, commissioning, shipping, etc. are significantly improved.

Stock optimization often does not have enough attention in a real environment. According to the detailed analysis presented by Bajrić [1], inventory management in the average company from Bosnia and Herzegovina is far from satisfactory. There are either too many products in the stock, so there is an unnecessary cost of keeping them, or not enough products, so there is a lost sales, cost of stopping production, replanting, switching to other products, breaking deadlines, returning to production of the original product and related costs.

According to the mentioned research, stocks in the average Bosnian company can be reduced by an average of 25%. On the other hand, the average level of availability can be increased from the current 80%, while respecting the cost-effectiveness limits, to 90%. These indicators point to the fact that with sophisticated and modern algorithmic solutions, stocks can be significantly reduced while reducing cost of lost sales as well. On the other hand, the satisfaction of end customers is significantly improved by introducing such mechanisms, and the level of delivery and service is almost one hundred percent.

To explain this problem in more detail, let's use the example of a company with a turnover of 10 million EUR (€). Let's assume that the average inventory turnover ratio is 10, although many companies are not at that level. Let the profit margin be 8%, which is an industry average. According to the calculation shown in Figure 1, the cost for not having time to optimize stock, for this company is 17.8% of the profit annually.

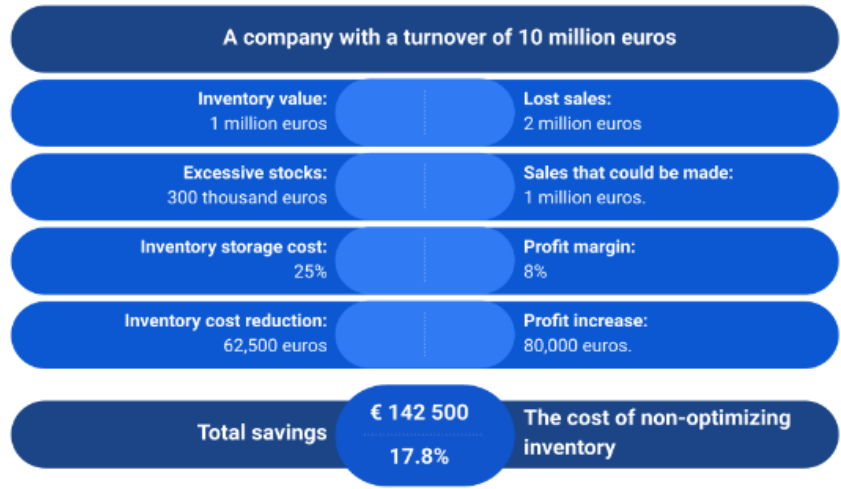

Figure 1. Real company analysis - percentage of lost sales.

Figure 2 presents a slightly different analysis. For a company with a lower profit margin than e.g. 5%, the loss is even higher, i.e. 22.5%, while a company that has a profit margin of 10% due to inadequate inventory management loses 16.3% of profit.

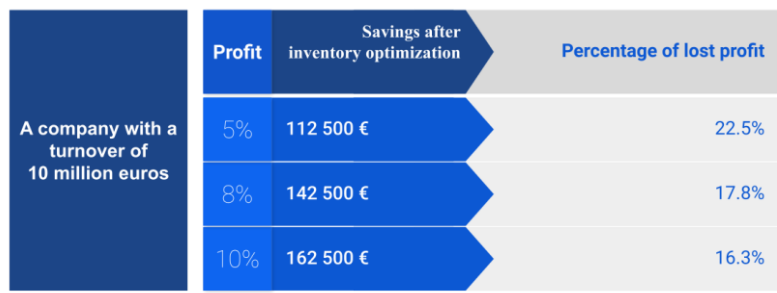

Figure 2. Real company analysis - percentage of lost profits.

These figures show analyzes for companies that in practice do not pay enough attention to quality inventory management.

According to the most relevant research, the annual cost of keeping stocks is from 20-50% of the value of stocks (Ma *et al.* [2]). This amount depends on the nature of the storage object itself, whether the product requires specific microclimatic conditions, how much storage space it occupies, how much storage space costs, how much the item loses value, how much it hardens, whether it is attractive for disposal, whether it is damaged, what is the opportunity cost of tied-up capital and other factors.

Lost sales, as one of the possible consequences of depletion of inventories, is typical in the stock management of finished products. The dominant cost is in the stock management of products that have short shelf life, such as e.g. bakery products, fresh fruits, vegetables, meat, daily newspapers, etc. The reason is, on the one hand, that the lost sale can not be proved exactly, while on the other hand, the expired goods that were returned from the point of sale are physically visible and represent the clearest material evidence.

Figure 3 shows the stock status of a typical FMCG (*fast-moving consumer goods*) product. Although it is not a product that has a short shelf life, very often there are lost sales due to depletion of stock. Although the stock of this item may be reduced, the dominant cost of this item is lost sales. In the covered period of 395 days, depletion was performed 84 times, which represents 21% of the days without stocks.

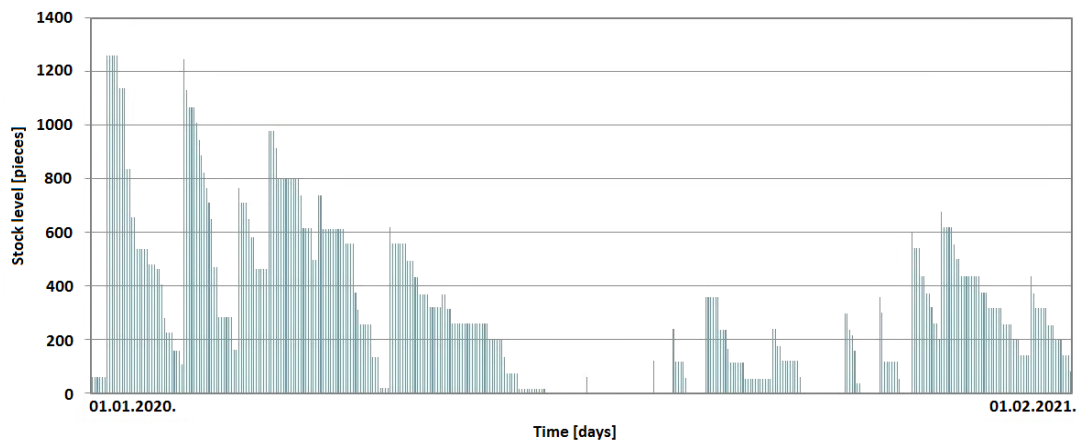

Figure 3. Real example of the state of stocks in a warehouse - lost sales.

Based on the above analyzes, it is easy to conclude that there is a huge potential in the real environment, which is mostly based on successful methods of sales prediction. On the other hand, if sales data are observed throughout history, then it can easily be reasoned that the data is in the so-called time-series format. There are many methods for forecasting based on time-series data, and this paper, as a continuation of the paper by Žunić *et al.* [3], presents a comparative analysis of three modern algorithms: Facebook's Prophet, Amazon's DeepAR+ and Amazon's CNN-QR (Convolutional Neural Network - Quantile Regression). The analysis is based on a comparison of the performance of the given algorithms depending on the available historical sales data, as well as on the length of the history itself for each of the analyzed items. All analyzes presented below were performed on real data obtained from one of the biggest distribution companies in Bosnia and Herzegovina. Dataset is available at the link [4], in order to be available to the rest of the researchers, as a new benchmark data.

The structure of this paper is as follows: section Literature review offers a general description of previous studies relating to the use of the different approaches and algorithms in related retail sales forecasting problems, and methods for solving. Section Case study gives an overview of the proposed forecasting approach by briefly explaining a structure of input dataset, data filtering and preprocessing steps, the process of product portfolio selection, the used algorithms (Prophet, DeepAR+ and CNN-QR), performance metrics and forecastability analysis using backtesting experiments, and guidelines for classifying the product portfolio, whereas section Results and discussions shows the capabilities of the proposed forecasting approach in a real-world use case scenario with significant comparison analysis. The conclusions drawn from the results in terms of the proposed objective are given in section Conclusions, with a brief description of directions for future work, further development and application of the proposed approach, with the focus on real-world usage.

## 2. LITERATURE REVIEW

Since the sales are dependent on many factors, the sale forecasting is not an easy job. To increase the accuracy of sales forecasting, Dwivedi *et al.* [5] use Adaptive Neuro Fuzzy Inference System (ANFIS) method. They forecasted the sales of the automobile industry in India based on forecasted sales produced by two forecasting methods moving average and exponential smoothing as input variables to the ANFIS method. They also compared ANFIS with artificial neural networks (ANN) and linear regression. At the end, they concluded that the ANFIS method gives more accurate results. Aburto and Weber [6] propose a hybrid intelligent system combining Autoregressive Integrated Moving Average (ARIMA) models and neural networks for demand forecasting, which helps to improve supply chain management in the retail industry. Presented forecasting model leaves valuable information for the respective business, e.g. they permit to quantify the effect of special events (such as holidays, end of month, etc.). Retail businesses are forced to use their resources efficiently and to make strategic decisions for the future in order to survive and increase their revenues. Comparative study on retail sales forecasting between single and combination methods by Aras *et al.* [7] shows that the combination methods achieve better results than the individual ones.

Alon *et al.* [8] made a comparison between ANN and traditional methods including Winters exponential smoothing, Box-Jenkins ARIMA model, and multivariate regression. Results showed that the ANN models were the best, followed by the Box-Jenkins model. Kang [9] in his Ph.D. dissertation concluded that the ARIMA model has a superior or equivalent mean absolute percentage error (MAPE) than the ANNs. The forecast error for the ANNs is lower when trend and seasonal patterns are in the data. Ansuj *et al.* [10] analyzed the behavior of sales in a medium size enterprise for the 10 years period of history data. Authors found that the forecasts obtained using the backpropagation model were more accurate than those of the ARIMA model with interventions. Kolassa [11] considered discrete predictive distributions to forecast daily retail sales and explained why forecast accuracy measures are inappropriate for count data. Jiménez *et al.* [12] wanted to obtain more accurate forecasts for online sales and to find out the relevant features of the sold products that affect the sales, so he proposed a selection methodology of a novel feature. Papacharalampous and Tyralis [13] assess the performance of random forests and Prophet in forecasting daily streamflow up to seven days ahead in a river in the US based on past streamflow observations. Random forests additionally used past precipitation information. Authors also implemented a naive method, as well as a multiple linear regression model utilizing the same information as random forests. The obtained results suggest that random forests perform better in general terms, while Prophet outperforms the naïve method for forecast horizons longer than three days.

Salinas *et al.* [14] propose DeepAR methodology for producing accurate probabilistic forecasts, based on training an autoregressive recurrent neural network, which learns a global model from historical data of all time series in the dataset. Their model can handle widely-varying scales through rescaling and velocity-based sampling, generates calibrated probabilistic forecasts with high accuracy

For the successful operation of distribution companies, it is necessary to ensure the efficient execution of all standard processes of the Supply Management Systems (SCM): ordering goods from manufacturers, warehousing and order picking, as well as the transport to end customers. The first segment is based on the predictions of inventory and sales, which provides warehouses that can meet all customer needs, and do not accumulate unnecessary goods and thus do not increase costs. Warehousing is a complex set of processes managed by the Warehouse Management System (WMS). In [3], the concept of smart WMS is described. Zunic *et al.* [15] described the application of Facebook's Prophet algorithm for sales prediction as part of the smart WMS concept and optimization of distribution companies. The concept of smart WMS and sales prediction have been tested in real environments and over real data in some of the largest warehouses in Bosnia and Herzegovina.

The smart WMS contains a number of predictive processes, such as moving goods from the stock to the picking zone based on sales forecasts, described by Zunic *et al.* [16]. Moving goods to the picking zone of the warehouse in combination with quality positioning of goods in the warehouse provides quality prerequisites for the process of order picking (Zunic *et al.* [17]).

As a third segment of the SCM, it is necessary to implement an efficient Vehicle Routing Problem (VRP) solver. A multiphase approach improved by the GPS data analysis is described in two papers by Žunić *et al.* [18, 19]. Several machine learning models have been used to predict unloading times for each customer (Žunić *et al.* [20]) as well as other parameters of practical vehicle routing problems with realistic constraints (Zunic *et al.* [21]). By analyzing the collected data, it is possible to improve other processes in the distribution companies workflow. For the successful implementation of those improvements, it is necessary to ensure the accuracy of the data. An algorithm inspired by the QRS signal detection has been implemented to detect anomalies in global positioning system (GPS) data (Žunić *et al.* [22]) An approach based on the largest common divisor and median value for detecting anomalies in sales has been implemented as part of the SCM (Zunic *et al.* [23]) The whole transport optimization process in the real environment could be improved by using external data analysis, such as social networks data about traffic tracking is presented by Žunić *et al.* [24].

## 3. CASE STUDY

In this section of the paper, the methodology used will be described. The structure of the dataset, the method of data collection, filtering and preprocessing will be described. Real data collected through several years of sales of the distribution company in Bosnia and Herzegovina were used. Dataset is publicly available at the link [4] as a new benchmark dataset.

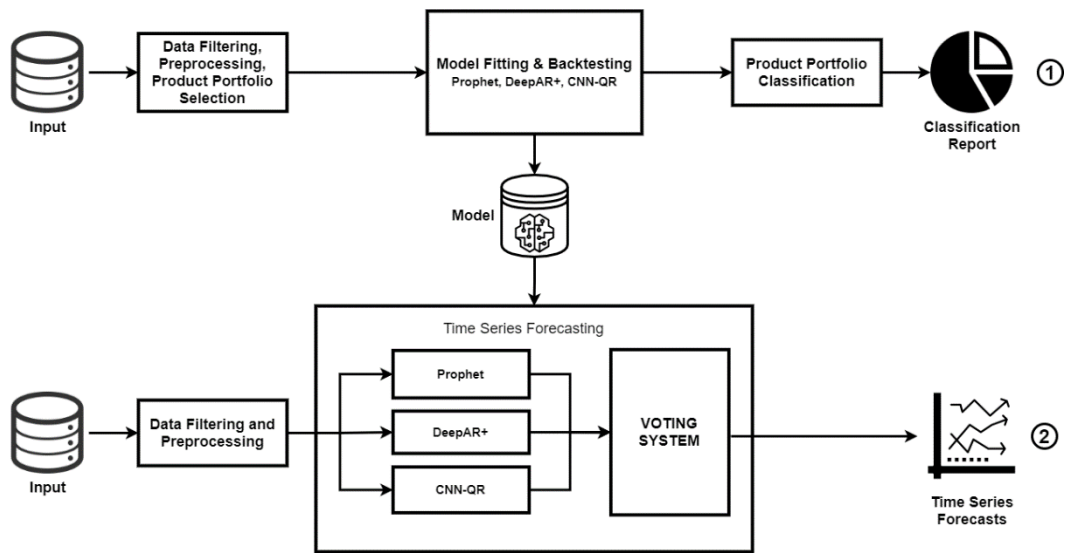

Figure 4. Proposed forecasting approach.

A comparison of three modern forecasting algorithms was made: Facebook's Prophet, and Amazon's DeepAR+ and CNN-QR algorithms. The parameters used, metics, and the method of backtesting are described. This lists all the steps for the successful implementation of these algorithms in practical use, and for their comparison. Figure 4 shows a sales forecasting approach inspired by the model described in [3], but based on the three mentioned algorithms.

### 3.1. Dataset, Data Filtering and Preprocessing

Sales data are observed. The data are aggregated on a monthly basis, due to the possibility of obtaining more accurate predictions and due to practical needs. Complete prediction is done at the item and organization level.

Amazon Web Services (AWS) have the ability to find connections between similar data. With the observed data structure, that possibility is increased. Therefore, the algorithm can improve the prediction results in one warehouse based on the prediction results of the same item in another warehouse. Facebook's Prophet forecasting algorithm observes only an article-organization pair without seeking for additional connections.

During the preparation of the dataset, data preprocessing and filtering was done. The last seven years of sales were observed when possible. Daily sales are aggregated, and the average monthly price is calculated for each pair of item and organization. In addition, a unit quantity was determined for each item, and an appropriate price was calculated. The price of an item can change over the months, and may depend on the observed organization.

The created dataset [4] contains data on quantities and prices of items. The data are divided into two groups, where the first contains the following columns:

- **item** - item identification number;
- **org** - organization identification number;
- **date** - the date of sale presented as the first day of the month;
- **quantity** - quantity of items sold.

The second group contains data on item prices. It contains the following information:

- **item** - item identification number;
- **org** - organization identification number;
- **date** - the date of sale presented as the first day of the month;
- **unit_price** - the price of an individual item.

An example of data in both groups is given in Table 1.

Table 1. Record format in the dataset.

| | item | Org | Date | quantity |
|---|---|---|---|---|
| **Quantity** target_ts.csv | 3959294 | 1617388 | 2021-01-01 | 3718 |
| | **item** | **Org** | **Date** | **unit_price** |
| **Price** related_ts.csv | 3959294 | 1617388 | 2021-01-01 | 0.611830413 |

## 3.2. Product Portfolio Selection

The analyzed dataset contains the 50 most significant items. Sales in four organizations were observed for each item. The observed data represent 99% of the sales of the analyzed brand. When determining the significance of an item, the total financial return after the sale per item is observed. For the implementation of backtesting, the minimum necessary data history is determined. Based on empirical analysis, it was concluded that 18 months of history is sufficient to observe the necessary changes in the data (seasonality and trend). Due to the need for additional data for testing, items with a history of at least 24 months were observed. Thus, the amount of necessary data in relation to the dataset described in [3] was reduced, and the number of items for testing was increased.

## 3.3. Algorithms

The paper provides a performance comparison of Facebook's Prophet algorithm, and Amazon's DeepAR+ and CNN-QR algorithms.

**The Prophet Forecasting Model** is an open-source procedure for forecasting time-series data. It was created by Facebook's Core Data Science team. The Prophet algorithm works well in the case of datasets with multiple seasons, and qualitatively describes the seasonality in the data.

Taylor and Letham [25] concluded that the Prophet Forecasting Model is designed to allow intuitive parameter adjustment without the necessary knowledge of underlying model details. To create the model, a decomposable time-series model is used (Harvey and Peters [26]). The model is based on the analysis of three components. It is based on an additive model where the first component is the trend of data behavior. Nonlinear data behavior is described by seasonality on a daily, weekly, or annual basis. The algorithm also observes the effects of holidays, which increases its accuracy.

The final estimate is formed on the basis of the expression:

$$y(t) = g(t) + s(t) + h(t) + \epsilon_t,$$

where $g(t)$ represents the trend, $s(t)$ represents the seasonality, while $h(t)$ represents the holiday effect on data behavior. The value of $\epsilon_t$ is an error and change in the data that is not contained in the model.

The proposed version of Prophet Forecasting Model uses a saturating growth model and a piecewise linear model for the trend analysis. Seasonality is based on the Fourier series, thus creating a flexible model for the detection and analysis of seasonality in the data (Harvey and Shephard [27]). Holidays significantly affect data behavior. Apart from the holidays that are repeated in the same period of the year, there are a large number of events that affect the behavior of the data, and there is no fixed date in the year. In addition, holidays vary by country and are often based on the lunar calendar. The Prophet model looks at global holidays and holidays by country, and assumes that events are independent. The final implementation is similar to the seasonality analysis.

The algorithm is open-source and is available in Python and R programming languages. It is used for a large number of forecasts within the Facebook platform, and based on the paper [25], it achieves extremely high quality results.

**The DeepAR+ model** is based on autoregressive recurrent neural networks [28]. It is used as a supervised learning algorithm to forecast one-dimensional time series. DeepAR+ has been trained on a large number of time series. A global model based on all the data in history has been created. The model has been proposed by researchers at Amazon. The training dataset includes a variety of items, with different lengths of history and sales status. The most important contributions of DeepAR+ are the use of recurrent neural network architecture for predictions, and empirical analysis of results over inputs with different characteristics, and obtaining quality results.

Unlike standard approaches, such as ARIMA (autoregressive integrated moving average) or ETS (exponential smoothing), one model was used to fit multiple time series. Therefore, the model has the ability to learn seasonality and dependence between different time series, with minimal modifications. It recognizes the complex relationships in the observed data. DeepAR+ learns based on a previously defined dataset, and for new data it can learn with a significantly shorter history, which is a great advantage over classical methods.

The DeepAR+ model is trained using random sampling, and recognizes the trend and seasonality. It achieves significantly better results for cases of datasets with a large number of features. The algorithm is not open source.

Amazon's **CNN-QR** (Convolutional Neural Network - Quantile Regression) forecasting algorithm is based on the application of casual convolutional neural networks to predict scalar time series data [29]. The algorithm creates one global model based on a large dataset, and uses a quantile decoder to create probabilistic predictions. The algorithm is not open source.

The CNN-QR algorithm, similar to DeepAR+, has the ability to find related time series, and spot seasonality and other important components in behavior. It is based on checking the quality of the reconstruction of the decoded strings in relation to the encoded data strings. Quantile regression is used with convolutional neural networks to extract features.

Amazon's algorithms are only available as part of AWS forecast services. For the testing process, default parameter values were used. For Facebook's Prophet algorithm, modified parameters based on the human analyst described in [3] were used. Parameter tuning was not done on the dataset used for testing, so no overfitting occurs.

Algorithms use price as a regressor. Information on local holidays in Bosnia and Herzegovina is used. Implemented algorithms are used for predictions on a monthly or quarterly basis. Therefore, the proposed analyzes will contain such predictions. All the above algorithms can be used for daily, weekly, monthly or annual predictions.

### 3.4. Performance Metrics & Backtesting

To measure the accuracy of predictions, **WAPE** (Weighted Average Percentage Error) was used. This metric was chosen in relation to the standard MAPE (Mean Absolute Percentage Error) metric due to the occurrence of falsely high results for items where sales are close to zero in certain months.

To check the quality of the model, backtesting was used. For backtesting implementation, any moment in the data history is set as the current time. All earlier data is used as history to create the model, while newer data is used for testing. Using this testing model, it is possible to do testing without waiting for future events. As mentioned earlier, due to annual changes and seasonality, it is necessary to use at least 18 months of historical data.

International Journal of Computer Science & Information Technology (IJCSIT) Vol 13, No 2, April 2021

For backtesting on a monthly and quarterly basis, the following is done: for the first two months, only monthly errors are observed. For each of the following months, a monthly and quarterly error are calculated. Therefore, ideally, it takes 12 months to test. Monthly WAPE is calculated based on 12 samples, while quarterly WAPE is calculated based on 10 samples.

Instead of a fixed number of 12 backtesting steps, a minimum (6) and a maximum (12) number of steps are defined. Therefore, backtesting can be done for items that have an 18+6 months history. If possible, backtesting with 12 steps is done, while if not, backtesting is done for a possible number of steps. The results remain unchanged for items with a longer history (30+ months), while for items with a shorter history (24-29 months) a shortened backtesting is performed. This results in WAPEs that are less reliable but of sufficient quality for practical use.

Due to this testing scheme, testing was done in our test environment, regardless of the fact that the AWS Forecast service offers the possibility of backtesting and comparison with Prophet. Another reason is the fact that the AWS Forecast service sets a limit of a maximum of 5 backtesting steps, which contradicts the set assumption of 12 backtesting steps if possible.

## 4. RESULTS AND DISCUSSIONS

The main goal of the paper is to provide a comparison of three forecasting algorithms: Prophet, DeepAR+ and CNN-QR. Due to the different approach in creating the mentioned models, a comparison was made in two phases. In the first phase, articles with a longer history were observed, while in the second phase, algorithms for articles with a shorter history were compared. Items are said to have a longer history if the dataset contains at least 24 months of sales data.

The first phase aims to test the possibility of improving the application of Facebook's Prophet algorithm described in [3]. In the second phase of testing, the goal is to expand the set of observed items with items with a shorter history. Thus, a possible combination of the proposed methods would achieve a quality result for a wide range of items.

Figure 5 shows the distribution of items sorted by importance. Significance is observed in relation to the total financial turnover. The x-axis represents the importance index.

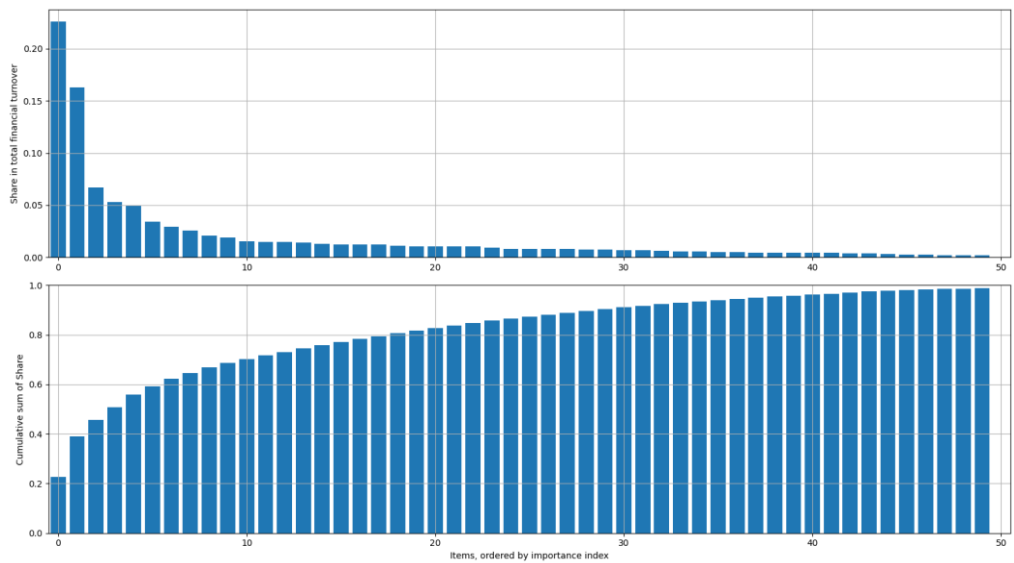

Figure 5. Distribution of items sorted by importance.

75

International Journal of Computer Science & Information Technology (IJCSIT) Vol 13, No 2, April 2021

At the same time, Figure 6 shows a chart that classifies items into active and inactive, and items with a short and long history.

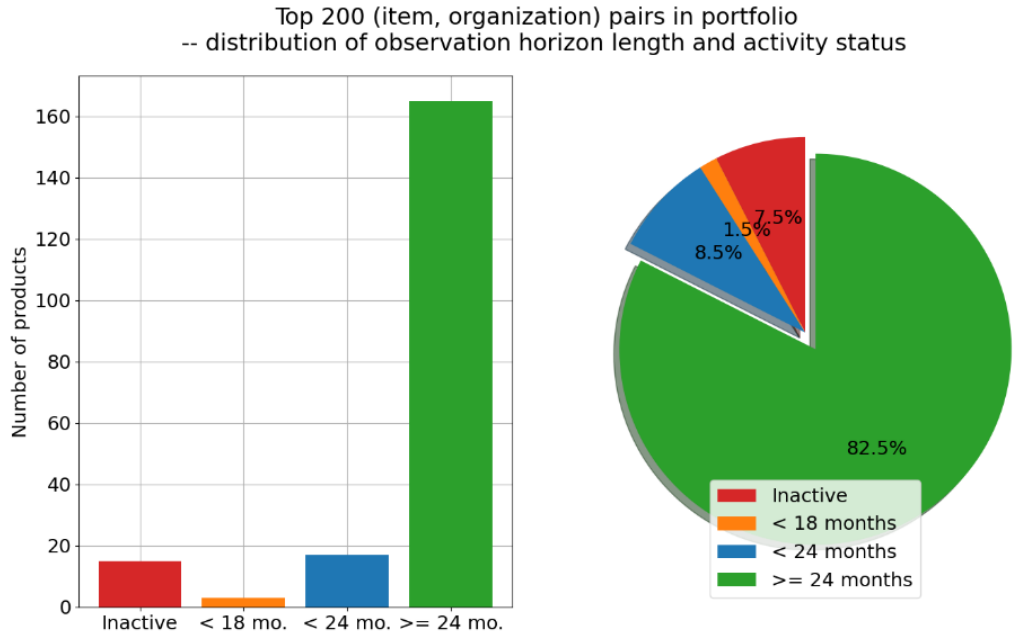

Figure 6. Graphic representation of active and inactive items with length of history.

By analyzing Figure 6 in more detail it can be concluded that more than 90% of items in the portfolio are active and have enough length of history.

### 4.1. Items with a Long History

In Figure 7, a visualization of accuracy for the financial importance of the item is created. A comparison is given for all three algorithms, with up to four points shown for each item on the x-axis (for each organization). The red line represents first-order interpolation, and represents a visualization of error growth by reducing the importance of the item. The line shows better quality results of the Prophet algorithm for important items, but also a decrease in performance by reducing the importance of items.

International Journal of Computer Science & Information Technology (IJCSIT) Vol 13, No 2, April 2021

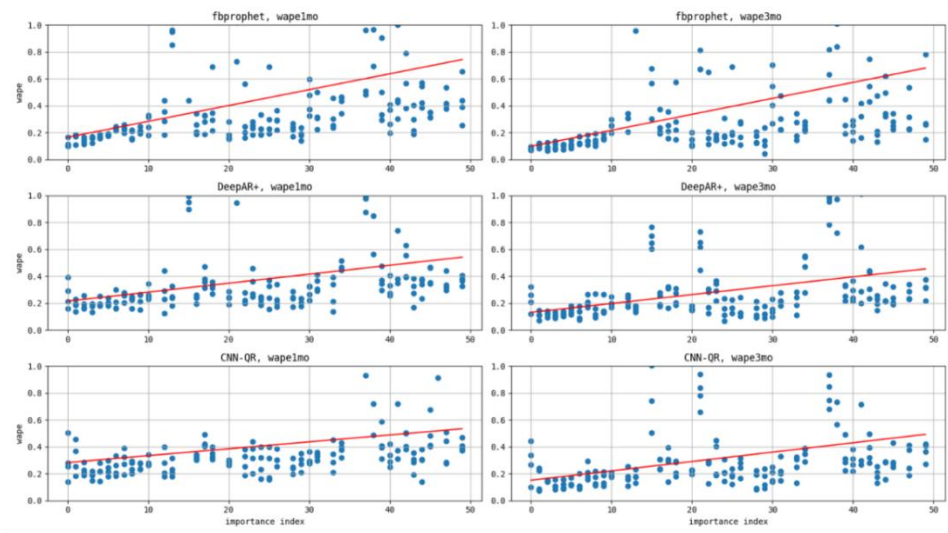

Figure 7. Graphical representation of the behavior of WAPE values by changing the importance index.

Figure 8 presents the data from Figure 7 in the form of cumulative histograms. The columns represent monthly and quarterly forecasts, while the rows represent the 10, 25, and 50 most important items with long history. Graphs can easily read the percentages of items for which successful predictions can be made.

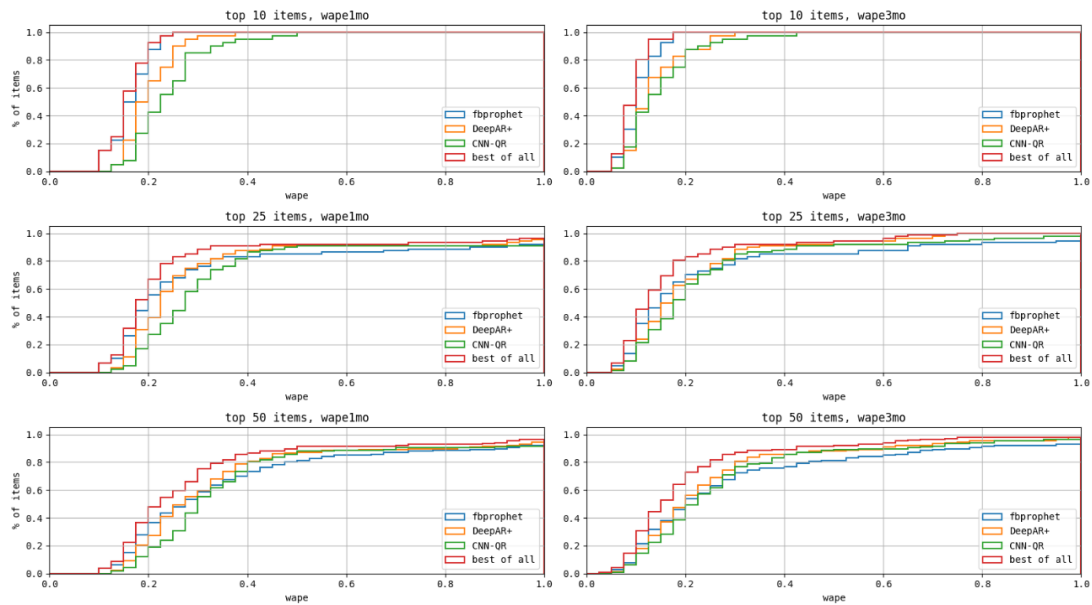

Figure 8. Cumulative histograms for predicting sales of the most important items.

Therefore, the left histogram primarily shows that on a monthly basis with an error less than WAPE = 0.2 Prophet can predict about 70%, DeepAR+ about 50%, and CNN-QR about 30% of the most important items.

Figure 9 shows the 5 most important items with a detailed comparison of WAPE for 1 month (wape1mo) and WAPE for 3 months (wape3mo), and for each item and organization. For the most

77

important items, Facebook's Prophet algorithm achieves dominant results, which is especially evident in the forecast for one month in advance. Data with a long sales history are presented.

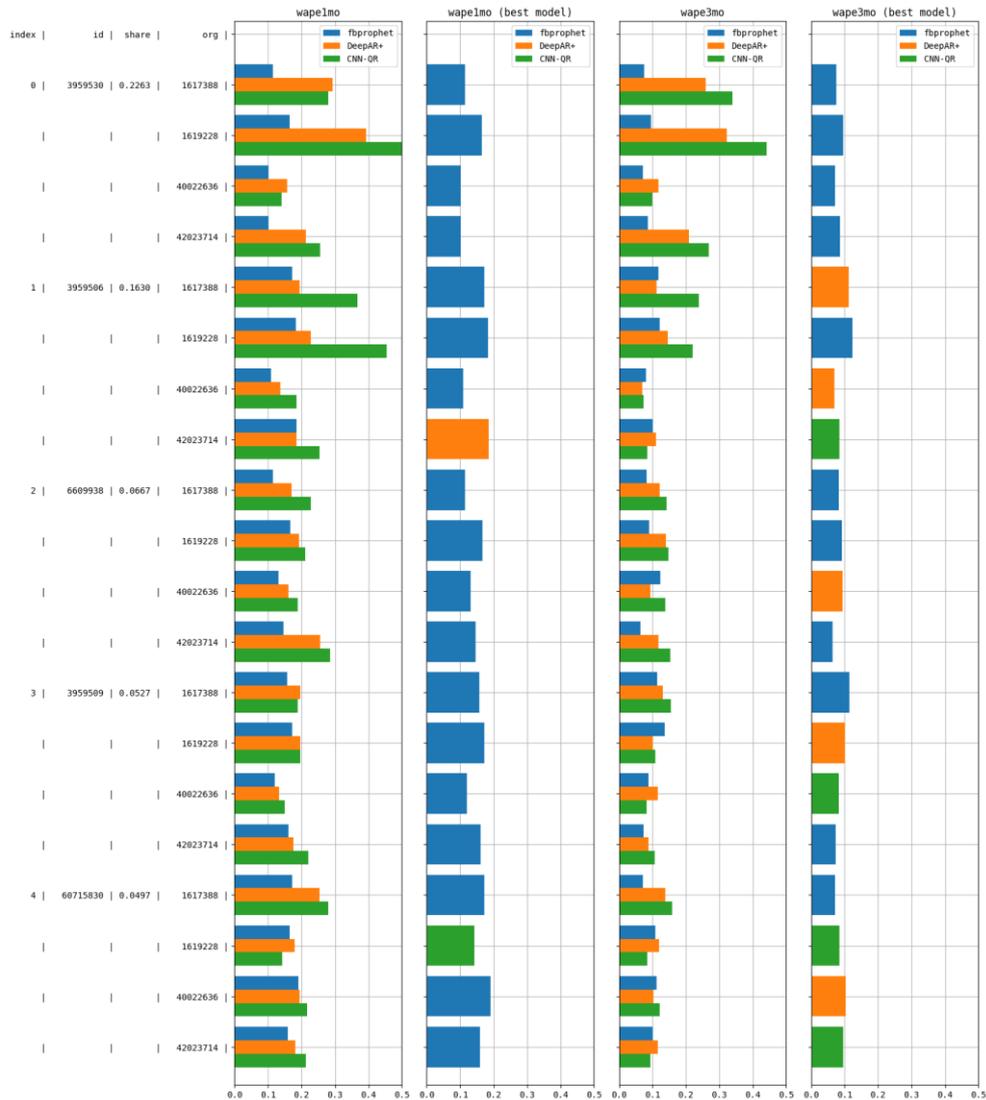

Figure 9. Detailed comparison for the top 5 items.

Figure 10 shows a stacked bar plot diagram. For each value of the importance index, the percentage of items for which the model is best is calculated. When comparing, only items from the corresponding importance index are observed. For example, if only the top 10 items are viewed on a monthly basis, in ~75% of cases the Prophet is the best model. In 15% of cases DeepAR+ is the best model, while in ~10% of cases CNN-QR is the best model. Analogously, if we look at the 50 most important items, Prophet is the best in 50% of cases, while the remaining two algorithms are the best in ~25% of cases.

International Journal of Computer Science & Information Technology (IJCSIT) Vol 13, No 2, April 2021

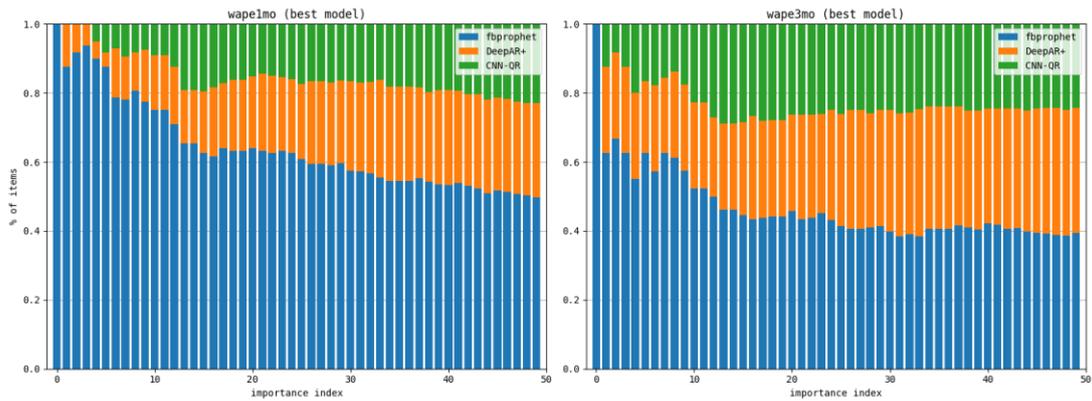

Figure 10. Representation of algorithms in the "best of all" model depending on the importance index.

These results confirm the dominance of Facebook's Prophet algorithm for the most important items. By reducing the importance of items, Prophet's dominance over other algorithms decreases, while for quarterly predictions and for the top 50 items, all algorithms are evenly represented in the "best of all" model.

### 4.2. Items with a Short History

Similar analyzes were performed for items with a shorter history. Due to the specifics of the items and the work of distribution companies, the number of such items is significantly smaller.

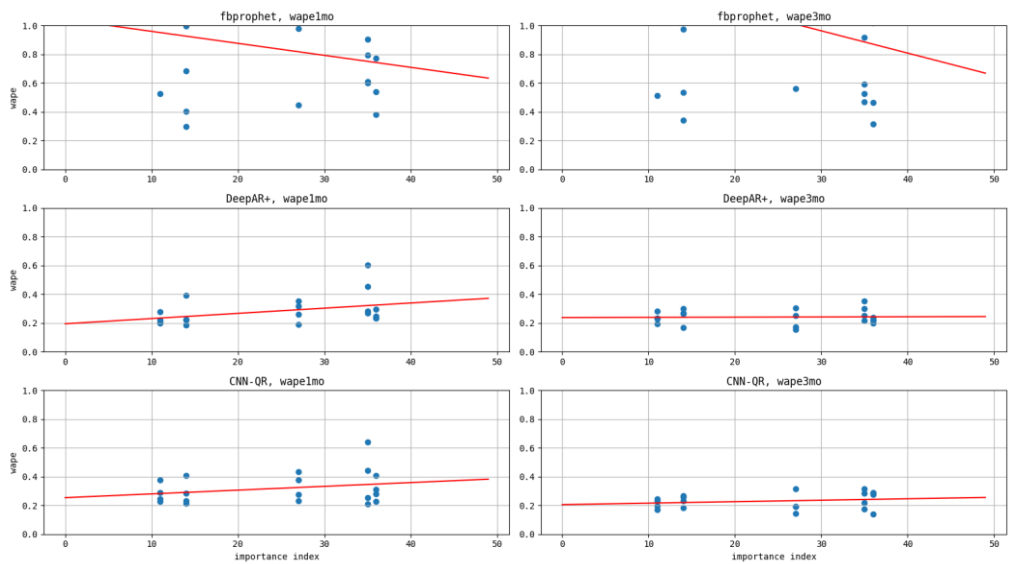

Figure 11. Graphical representation of the behavior of WAPE values by changing the importance index for articles with short history.

Given the consistency of the results, the described conclusions were drawn. As mentioned earlier, Amazon's algorithms create a unique model based on all the data, while Facebook's algorithm creates one model for each item and organization. Therefore, the superiority of DeepAR+ and CNN-QR algorithms in the analyzes is expected. The results confirm that the performance of Amazon's algorithms is comparable to the results for items with a longer history, which is a significant result.

International Journal of Computer Science & Information Technology (IJCSIT) Vol 13, No 2, April 2021

Figure 11 shows a scatter plot for items with a shorter sales history. At the same time, in Figure 12, a cumulative histogram for sales predictions is given. The obtained results confirm significantly better performance of Amazon's algorithms compared to Prophet.

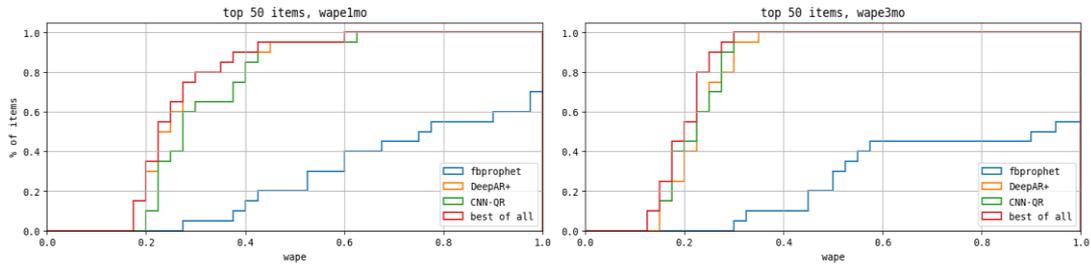

Figure 12. Cumulative histograms for predicting sales of items with short history.

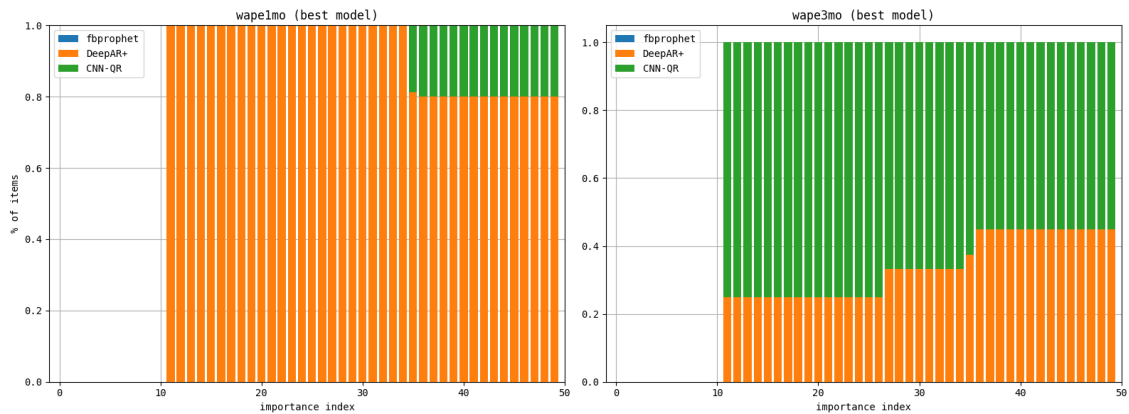

Figure 13. Representation of algorithms in the "best of all" model depending on the importance index for items with short history.

Finally, Figure 13 shows a stacked bar plot diagram for the performance of algorithms for items with a short history. The obtained results confirm that the success of Amazon's algorithms in relation to Facebook's Prophet is incomparable for articles where there is not a sufficient set of data for training per pair of articles and organizations.

Performance analysis is not crucial for practical application, because sales predictions do not have to be done in real time. In practice, the preparation of the model, as well as the predictions themselves, are often done outside working hours, which opens a large time window for creating predictions.

### 4.3. COVID-19 Analysis as a Forecasting Challenge

One of the important factors in the forecasting process is the ability to solve and cope in unexpected situations. One of them is the COVID-19 crisis, which significantly affected the work of distribution companies.

Figure 14 shows the analysis done in the period 12 months before COVID-19, and the backtesting analysis done for the first 12 months of COVID-19 in Bosnia and Herzegovina.

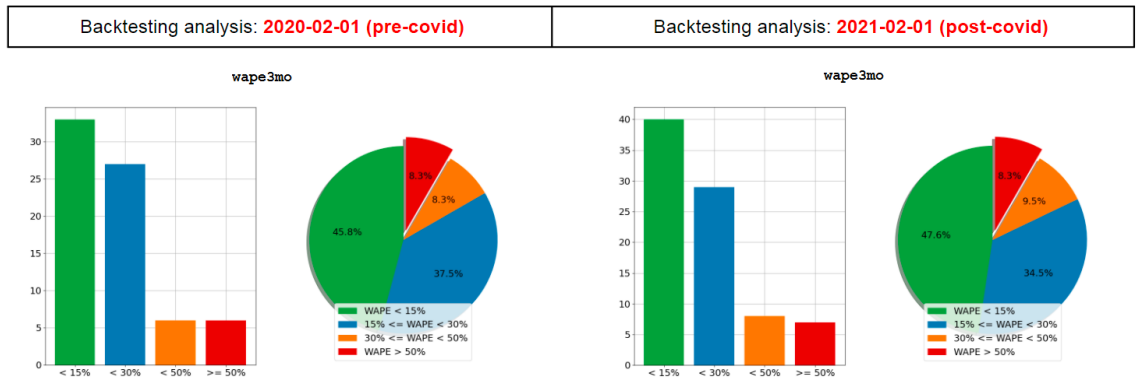

Figure 14. Analysis of forecasts caused by the COVID-19 crisis.

The performed backtesting analysis on the most popular items showed insignificant changes in the obtained results, which proves that the proposed forecasting model can cope well with unexpected changes.

Analyzing obtained results shown in Figure 14 in more detail it can be concluded that the forecasting is a bit better after the first 12 months of COVID-19 in Bosnia and Herzegovina, mostly caused by longer data history.

### 4.4. Discussion

The obtained results confirm the advantages of each algorithm. The question of the benefits of using each of the proposed approaches is raised, given the essential difference in the way Prophet works and Amazon's algorithms. As noted, Prophet analyzes each signal independently, while AWS algorithms create a single model for all signals and try to find interdependencies. Therefore, AWS algorithms show superiority over classical methods only when they have a large number of signals over which to create a model, and in the case of articles with a short history.

It has been observed that the Prophet model shows superiority in the case of items that are sold frequently, in large quantities and have a long sales history. At the same time, AWS algorithms showed dominance in other cases.

This leads us to the possibility of combining these methods to create sales predictions for a wide class of items. With the combined use of algorithms, it is possible to make predictions for items with frequent and infrequent sales, as well as for items with a long and short history, and in all cases to obtain satisfactory prediction results. Therefore, an improvement of the prediction system described in [3] based on the Prophet was obtained.

The described concept based on the use of the above algorithms for sales prediction can be of great benefit for application in the real world. The concept can deal with a diverse range of items, items with frequent and infrequent sales and a longer and shorter history, which is not the case for classic forecasting approaches.

On the basis of the COVID-19 analysis, it was noticed that the described approach successfully copes with unexpected changes, and that it quickly adapts to new situations.

## 5. CONCLUSIONS AND FUTURE WORK

The paper describes the application of three modern forecasting approaches: Facebook's Prophet, and Amazon's DeepAR+ and CNN-QR models. Algorithms were applied to sales prediction for distribution companies. A detailed comparison was made over various items, where the performance of algorithms over items with different sales history was analyzed.

The obtained results show in which situations it is better to use which of the listed algorithms. The results show that Facebook's Prophet is of better quality for predicting the sale of frequent items with a longer history. At the same time, due to the principle of operation of algorithms, where one model is created for all items, Amazon models show quality results for algorithms with a short history or a smaller amount of sales data.

The paper is of great importance because, to our knowledge, the comparison of the three above mentioned algorithms on actual sales data is not described in the literature. The paper provides a detailed methodology, uses real data and shows the application on a real example.

In the future, the application of the described model for inventory optimization is planned, as well as the analysis of forecasting results on the quantities of inventories in warehouses. The concept can also be used to detect anomalies and warn of errors in created orders, which can improve the complete concept of a smart warehouse.

In addition, it is planned to optimize the parameters used for algorithms based on the application domain. The implementation of other algorithms is also planned, and their comparison with the results obtained using the Facebook and Amazon methods.

## ACKNOWLEDGEMENTS

The authors want to thank the company "Info Studio d.o.o." from Sarajevo, Bosnia and Herzegovina, for making this research possible through funding and providing access to necessary data.

## AUTHORS

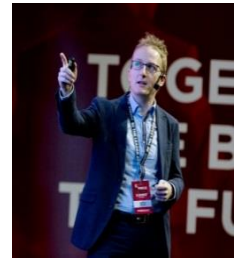

**Emir Žunić** is a PhD in Electrical Engineering with over 10 years of experience in the fields of Software Engineering, IT, Data Mining, Business Process Management, Document Management and Optimizations. He currently works as the Head of AI/ML Department at Info Studio d.o.o. Sarajevo. Also, he is the Co-Founder and CIO of edu720 d.o.o. Sarajevo. In the Academic Area, he also has experience in working as a Teaching Assistant/Industry Expert at the Faculty of Electrical Engineering, University of Sarajevo. In the past, he also worked as an Industry Expert at the Sarajevo School of Science and Technology. He has published more than 50 scientific papers at prestigious conferences and journals. He is an Editorial Board Member of several scientific conferences and journals.

**Kemal Korjenić** is a research engineer and data scientist at Info Studio d.o.o. Sarajevo. He gained his bachelor and master degree at the Faculty of Electrical Engineering, University of Sarajevo.

**Sead Delalić** is a Teaching Assistant at the Department of Mathematics at the University of Sarajevo and a research engineer at Info Studio d.o.o. where he is working on optimization projects and AI solutions in various practical domains. He gained his bachelor and master degree in mathematics and software engineering at the Faculty of Science. He is a receiver of two Golden Badges from the University of Sarajevo as one of the best students with an average grade of 10.00 (best possible). He is a PhD candidate at the Faculty of Science. His PhD focuses on nature-inspired metaheuristic algorithms.

**Zlatko Šubara** is a research engineer and data scientist at Info Studio d.o.o. Sarajevo. He gained his bachelor and master degree at the Faculty of Electrical Engineering, University of Sarajevo.